# The Ripple Pond: Enabling Spiking Networks to See

S. Afshar[1,2], G. Cohen[1], R. Wang[1], A. van Schaik[1], J. Tapson[1], T. Lehmann[2], T.J. Hamilton*[1,2]

[1]Bioelectronics and Neurosciences, The MARCS Institute, University of Western Sydney, Penrith NSW Australia
[2]School of Electrical Engineering and Telecommunications, The University of New South Wales, Sydney NSW Australia

Correspondence:
Tara Julia Hamilton
The MARCS Institute
Bioelectronics and Neuroscience
University of Western Sydney
Locked Bag 1797
Penrith NSW 2751
Australia
t.hamilton@uws.edu.au

*Abstract*—In this paper we present the biologically inspired *Ripple Pond Network* (RPN), a simply connected spiking neural network that, operating together with recently proposed PolyChronous Networks (PCN), enables rapid, unsupervised, scale and rotation invariant object recognition using efficient spatio-temporal spike coding. The RPN has been developed as a hardware solution linking previously implemented neuromorphic vision and memory structures capable of delivering end-to-end high-speed, low-power and low-resolution recognition for mobile and autonomous applications where slow, highly sophisticated and power hungry signal processing solutions are ineffective. Key aspects in the proposed approach include utilising the spatial properties of physically embedded neural networks and propagating waves of activity therein for information processing, using dimensional collapse of imagery information into amenable temporal patterns and the use of asynchronous frames for information binding.

**INTRODUCTION**

How did the earliest predators achieve the ability to recognize their prey regardless of their relative position and orientation? (Figure 1). What minimal neural networks could possibly achieve the task of real-time view invariant recognition, which is so ubiquitous in animals (Caley, M.J., 2003), even those with miniscule nervous systems (Van der Velden, 2008), (Tricarico, 2011), (Neri, 2012), (Avargues-Weber, 2010), (Gherardi, 2012), yet so difficult for artificial systems? (Pinto, 2008). If realised such a minimal solution would be ideal for today's autonomous imaging sensor networks, wireless phones, and other embedded vision systems which are coming up against the same constraints of limited size, power and real-time operation as those earliest animals of the Cambrian oceans

In this paper we outline such a simple network. We call this system the Ripple Pond Network (RPN). The name Ripple Pond alludes to the network's fluid-like rippling operation. The RPN is a feed-forward time-delay spiking neural network with static unidirectional connectivity. The network converts centered input images received from an unspecified upstream salience detector into spatio-temporal spike patterns that can be learnt and recognized easily by a downstream distributed Polychronous Network (PCN). Working together the entire system can perform shift, scale and rotation invariant recognition.

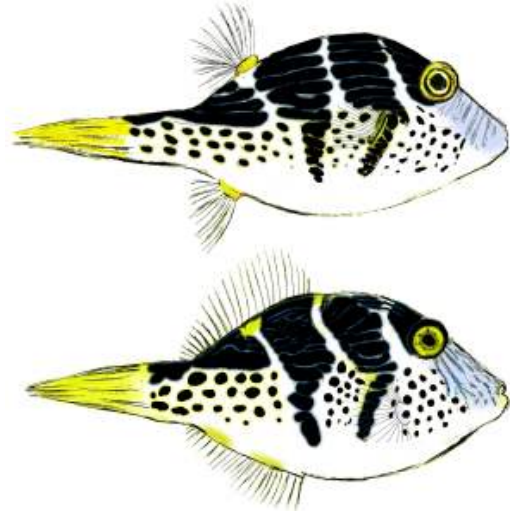

Figure 1. Batesian mimicry: The highly poisonous pufferfish, *Canthigaster valentine* (top) and its edible mimic *Paraluteres prionurus* (bottom). The remarkable degree of precision in the deception reveals the sophisticated recognition capabilities of local predatory fish whose neural networks, despite being orders of magnitude smaller than that of primates nonetheless seem capable of matching human vision in performance (note: the dorsal, anal and pectoral fins are virtually invisible in the animals' natural environment).

**The View Invariance Problem**
In the 2005 book, "23 Problems in Systems Neuroscience" (van Hemmen, 2005), the 16th problem in systems neuroscience, as posed by Laurenz Wiskott, is the view invariance problem. The problem arises from the fact that any real world 3D object can produce an infinite set of possible projections on to the 2D retina. Leaving aside factors such as partial images, occlusions, shadows and lighting variations, the problem comes down to the shift, rotation, scale and skew variations. How then, with no central control could a network like the brain learn, store, classify and recognize in real time the multitude of relevant objects in its environment?

Generally most biologically based object recognition solutions have been based on vertebrate vision, in particular mammalian vision, and have used either statistical methods (Sonka, 2007), (Gong, 2012), (Sountsov, 2011), signal processing techniques (such as log-polar filters) (Cavanagh, 1978), (Reitboeck, 1984), artificial neural networks (i.e. non-spiking neural networks) (Norouzi 2009), (Nakamura, 2002), (Iftekharuddin, 2011), and more recently, spiking neural networks (SNNs) (Meng 2011), (Serrano-Gotarredona, 2009), (Rasche, 2007), (Serre, 2005).

Since many of the approaches above are based on mammalian vision and achieving the accuracy and resolution of mammalian vision, they are very complex and can only be truly implemented on computers (Sonka, 2007), (Gong, 2012), (Norouzi 2009), (Nakamura, 2002), (Iftekharuddin, 2011), (Meng 2011), (Serre, 2005), (Jhuang, 2007), sometimes with very slow computation times. Other implementations that have been demonstrated on hardware (Serrano-Gotarredona, 2009), (Rasche, 2007), (Folowosele, 2011) have been successful in proving that vision can be achieved for small, low-power robots, UAVs, and remote sensing applications, however; the functionality of such neuromorphic systems has been limited.

The few models of invertebrate visual recognition have had an explanatory focus (Huerta, 2009), (Arena, 2012) and, not being developed for the purposes of hardware implementability, assume highly connected networks not suitable for hardware.

CMOS implementations of bio-inspired radial image sensors are most closely related to the RPN however, these sensors ultimately interface with conventional processors (Pardo, 1998), (Traver, 2010) rather than biologically plausible SNNs as is the case with the RPN. Figure 2 shows the human fovea (a), a typical log-polar implementation used to model this (b), the visual field (c) and mapping of this in V1 in a primate (d). While the mapping may look approximately log-polar in (d), the mapping at the fovea is approximately uniform which, together with the highly intricate connectivity required, reduces the biological plausibility of such log polar approaches.

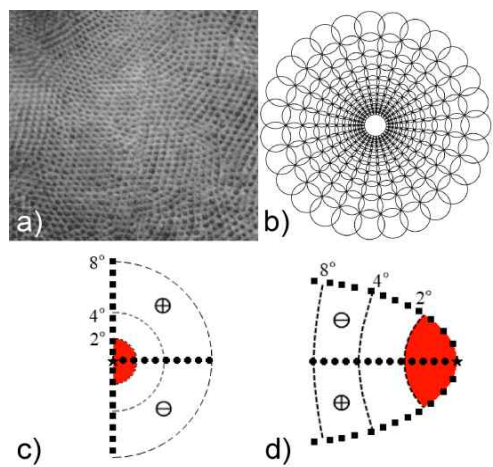

Figure 2. a) Human fovea centralis responsible for the central 2° of the visual field (two thumb widths at arm's length) (Ahnelt, 1987), b) Standard log polar model used in artificial vision (Traver, 2010), c) Visual field, and d) Visuotopic organization of V1 (i.e. the mapping of the visual field) in the monkey from (Gattass, 2005).

The RPN trades the high accuracy and resolution approaches favoured by other implementations, for speed and view invariance. Recent papers (Afshar, 2012), (Hamilton, 2011) showing the ability and efficiency with which SNNs, with competition between neurons, overcome problems with noise, incomplete data, power consumption and computation speed suggest that this is a good starting point to build a network capable of solving the view invariance problem. In the following sections we describe the learning network and the RPN that build on these principles.

**Polychronous Learning Network**
Before discussing the RPN, we will briefly describe the neuromorphic learning network with which it interfaces, namely: the PolyChronous Network (PCN) (Izhikevich, 2006). A PCN uses a spatio-temporal pattern of spiking neurons to represent information asynchronously whereas classic artificial neural networks discard timing information by modelling neuron firing rates sampled synchronously by a central clock. The PCN's additional use of asynchronous temporal information results in greater energy efficiency and speed (Levy, 1996), (Van Rullen, 2001). The PCN shares this feature with other recently proposed dynamic neural networks including liquid and echo state machines (Maass, 2002), (Jaeger H., 2001), however unlike PCNs, liquid state machines do not perform well for spatio temporal pattern recognition unless the reservoir network has been set up to ensure feedback stability, fading memory, and input separability (Manevitz, L., 2010), and as such we prefer to use the simpler, feed forward PCN for pattern recognition.

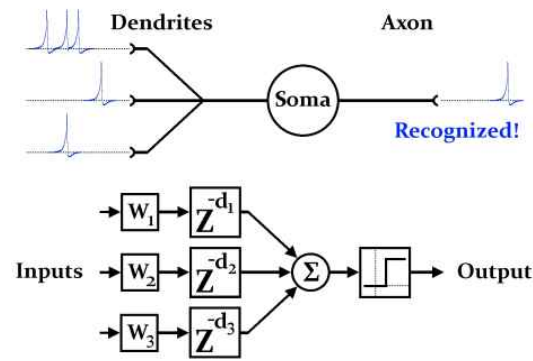

Figure 3. Biological model (top) and mathematical description (bottom) of a single element in a PCN.

A PCN can function as a content addressable, distributed memory system comprising of multiple neurons connected to each other via multiple pathways. The PCN through dynamic adaptation of synaptic weights (W1, W2, W3 in Figure **3**) and dendritic propagation delays (d1, d2, d3 in Figure **3**) makes particular neurons exclusively responsible for particular inter-spiking intervals. It achieves this by continuously adapting its parameters to maximize recognition at its output in response to the statistics of its input. Figure 4 shows an example of a spike pattern entered into a PCN with five neurons (Wang, 2011). Here in order for neuron 3 to spike, a spike at neuron 1 is delayed by T1+T2 and a spike at neuron 5 is delayed by T2. Thus the spikes at neuron 1 and 5 arrive simultaneously at neuron 3, causing it to spike. A longer spatio-temporal pattern can be stored in a network of neurons by learning the delays and weights that are required between neurons (Paugam-Moisy, H., 2008), (Ranhel, J., 2012). Subsequently the PCN's "output" can be measured as the relative activation of certain neurons which individually or in concert indicate the recognition of a certain pattern (Martinez, R., 2009).

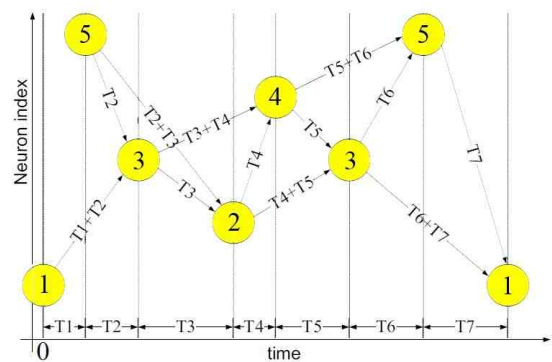

Figure 4. Neuron activation in a polychronous network (PCN) (Wang, 2011).

However PCNs or other distributed memory systems tasked with recognition cannot directly interface with the retina since they expect their learnt patterns to appear via the same channels every time (see Figure **5**). The RPN is in the class of systems that serve as the interface between such a memory system and retinotopically mapped inputs from the visual field. It does so by converting the raw visual stimulus to a Temporal Pattern (TP) suitable for the PCN, in this way the RPN is a transformation of an image from its 2-dimensional geometry into a 1-dimensional temporal sequence. Subsequent extensions of/to the RPN allow extraction of different image features which are presented to the PCN as additional TPs thus converting a two dimensional image to an m-dimensional spatio-temporal sequence.

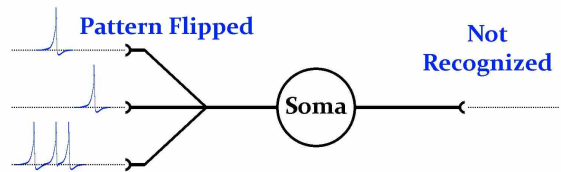

Figure 5.  A PCN element with the spiking pattern at the input flipped, compared with Figure 3, results in no recognition.

**MATERIALS AND METHODS**

**The RPN Network**
A central feature of the RPN is getting the most functionality from a minimally connected network.   Connectivity is a limiting factor which, though largely absent in biology, frequently constrains hardware implementations and yet is not often considered in the development of artificial neural networks (Sivilotti, 1991), (Hall, 2004), (Furber, 2012), (Park, 2012).

*An Upstream Shift Invariant Salience Network*
As shown in Figure **6**, the RPN receives an image as the spatio-temporal, high-pass filtered activation pattern of neurons on a conceptual 2D sheet representing the field of attention that has been produced by an upstream salience detection system. By using a sliding the window of attention and focusing it onto a single salient object at a time the salience detector effectively allows the overall network to operate in a shift invariant manner. The field of computational and biologically-based salience detection is extensive with a wide range of models and techniques and approaches (Drazen, 2011), (Gao, 2009), (Itti, 2001), (Vogelstein, 2007). The proposed RPN system does not require any specific salience model having a centered input image as its only requirement. For simplicity, however, we may assume the upstream salience network to consist of only a motion detector, which physically fixes the creature's gaze onto a moving object. In fact, this simplified system is not far off the mark in the case of many organisms (Land, 1999), (Dill, 1993) and may serve in robotics applications where energy and hardware are also limiting factors.

*Input Images Ripple Outwards on the Neuron Disc*
After centring by the salience network and high-pass filtering, the incoming "image" stimulates in a frame-based fashion the neurons distributed on a disk as shown in Figure **6**. For visual clarity in all the illustrated figures the disc comprises of only ten neurons on each arm ($N=10$) with a total of thirty arms ($\Phi=30$).

For $n=1…N$
For $\varphi=1…\Phi$

$$a_{(n+1),\varphi}(t) = a_{n,\varphi}(t) \times \delta(t-n) \quad (1)$$

where $a_{n,\varphi}(t)$ is the activity on the $n^{th}$ neuron on arm $\varphi$ at time $t$.

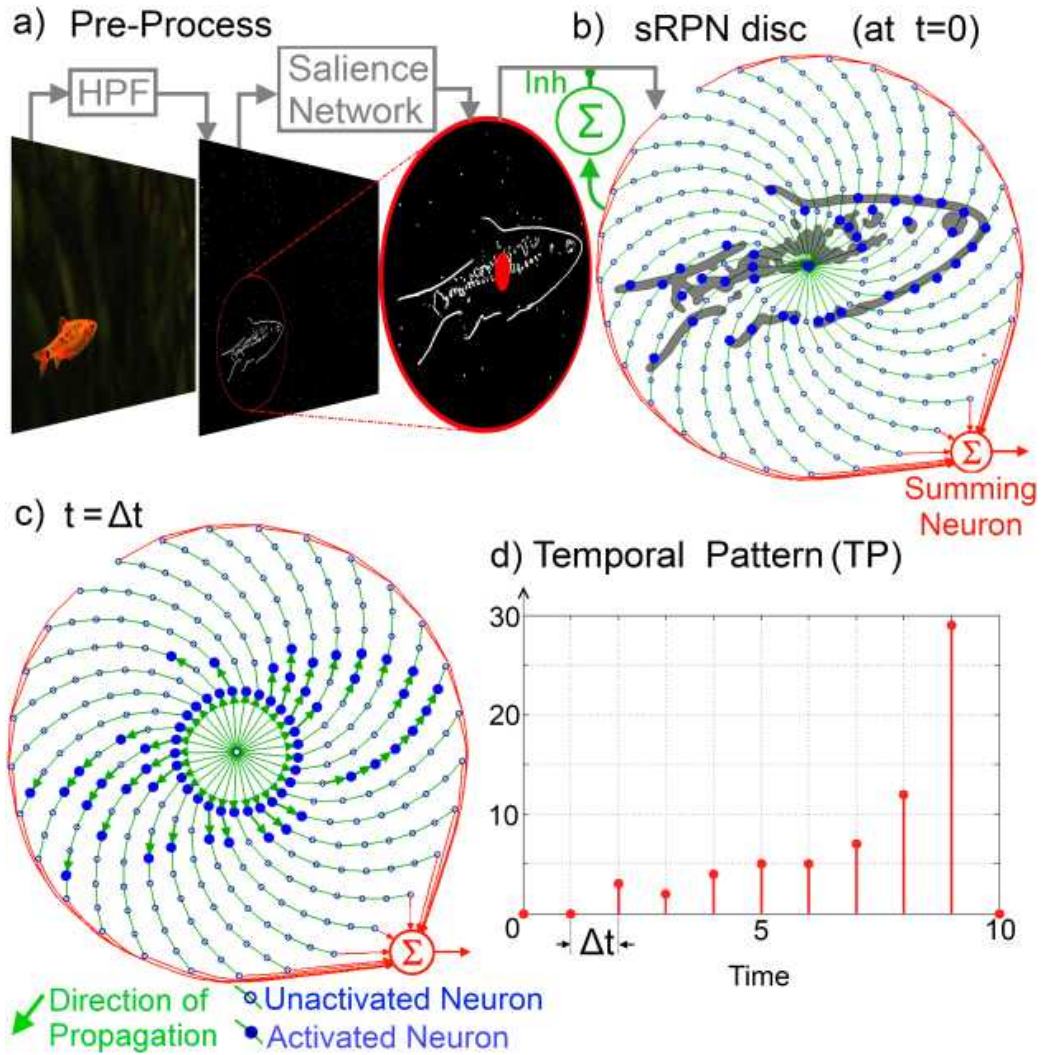

Figure 6. The spiral RPN System Diagram: Input image to Temporal Pattern generation.

Functionally these neurons represent simple binary relays with fixed unit delays between them. More complex neuron models (Indiveri, 2011) could increase the system's resolution but would incur an increased hardware cost. In the RPN each neuron $a_{n+1,\varphi}(t)$ can be activated if its upstream neuron (on the same spiral arm but closer to the disc centre), $a_{n,\varphi}(t)$ transmits a pulse to its input. All neurons are also sensitive to any incoming image, functioning as radially distributed outwardly connected pixels on a circular retina. With implementation in mind, and in contrast to log-polar filters, the density of the neurons along each arm of the disc is increased along the arm as we move away from the centre of the disc, permitting an approximately uniform distribution of neurons.

Starting from the disc's centre, the neuronal connections radiate outward to the edge of the disc. This outward connection creates a rippling effect. Since the neurons act as delayed relays, stimulation of a neuron close to the centre of the disc travels outward along the disc arm as a discrete wave, stimulating succeeding neurons in turn (1).

*The Summing Neuron outputs a Temporal Pattern*
The neurons at the edge of the disc all connect to a common Summing Neuron (red sigma in Figure **6** that outputs a real-valued TP that can be represented as a 1×N vector (2). From the geometry of the neuron disc it is clear that this output TP is rotationally invariant. More subtly but as important is the RPN's central feature of image-to-collapsed-temporal-pattern conversion. This conversion greatly simplifies the task of scale invariant recognition of the downstream PCN.

$$TP_{sum}(t) = \sum_{\varphi=1}^{\Phi} a_{N,\varphi}(t) \times \delta(t-n) \qquad (2)$$

Where:
$TP_{sum}(t)$ is the temporal pattern output of the summing neuron,
$\Phi$ is the total number of neurons on the disc,
$N$ is the total number of neurons per arm,
$a_{n,\varphi}(t)$ is the activity on the $n$th neuron on arm $\varphi$ at time $t$.

The summing neuron sums the activity of the neurons at the edge of the disc. Note that for simplicity we represent the summing neuron as a single neuron with a real valued spike output, but may also be represented by a network of neurons, which receiving the same input collectively produce the same output. The PCN however receives spatio-temporal spike patterns as inputs which are, to the first approximation, binary pulses. Thus, the real valued output of the RPN's summing neuron needs to be encoded into binary values via multiple channels into the PCN. This can be achieved in the PCN's input layer through the use of multiple heterogeneous neurons with a range of different internal parameters such as varied input weights, thresholds and center frequencies that would relay the value of the RPN output to the PCN's reservoir. These neurons would effectively act as multiple input channels which, though all receiving the same input, can provide an arbitrary number of channels into the PCN in a simple fan-out fashion.

*The Inhibitory Neuron Acts as an Asynchronous Shutter*
In addition to being sensitive to an incoming image and synapsing onto outer neurons along the disc arm, all neurons on the disc also connect to an inhibitory neuron (green sigma in Figure **6**) such that the inhibitory neuron carries information about the net activity of *all* neurons on the disc:

$$Inh(t) = \sum_{\varphi=1}^{\Phi} \sum_{n=1}^{N} a_{n,\varphi}(t) \qquad (3)$$

Where:
$Inh(t)$ is the output of the inhibitory neuron.

The output of the inhibitory neuron then feeds back inhibitively onto the visual pathway carrying the image to the disc. The Inhibitory neuron blocks this pathway preventing further transmission of the image. In this way the neuron ensures that as long as any activity remains on the disc (i.e. while RPN is processing the image) no new image will be projected onto it. In hardware this feedback loop can be realised via an asynchronous sample and hold; however, biologically plausible systems require a more distributed approach (Wang, 2010), (McDonnell, 2012), (Brunel, 2000) which we will consider further in the discussion section.

*Need for Normalization Depends on PCN Robustness*
In this paper we do not make assumptions about the functional complexity of the PCN and propose a complete system compatible with the simplest PCN realization; however we also suggest useful PCN functionalities that would enhance the RPN-PCN system. One example is the ability of the PCN to recognize spatio-temporal patterns received at different speeds which we here refer to as being robust. In order to compensate for a PCN lacking this capability we include the following network augmentation.

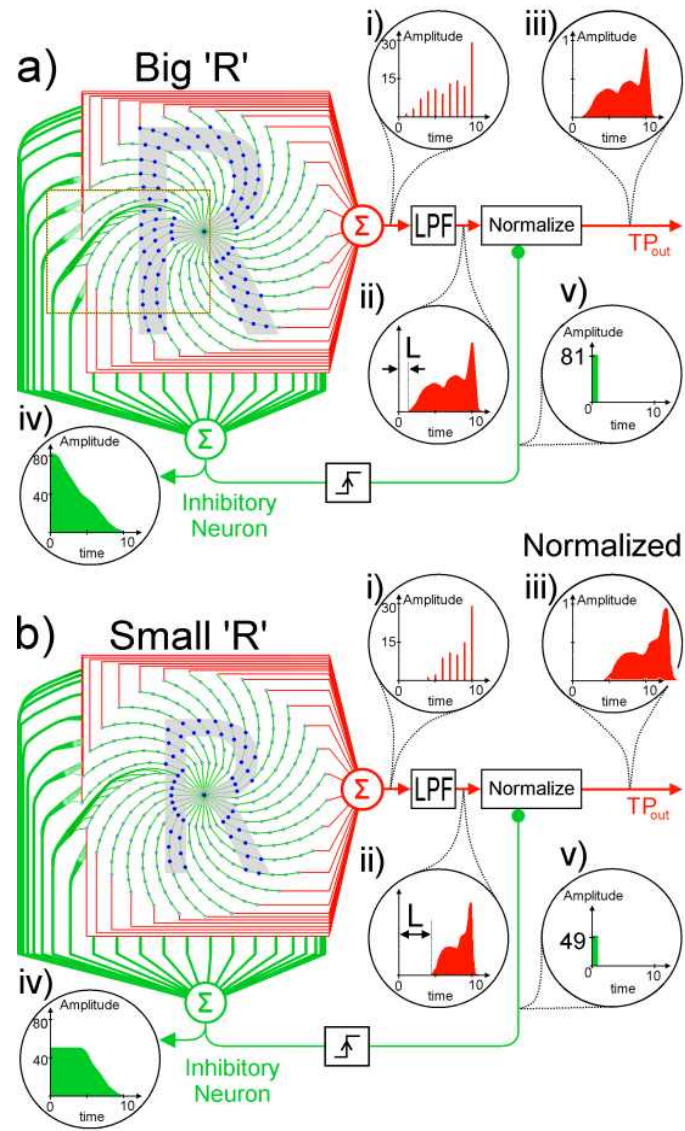

Figure 7. Normalization of amplitude and time of the TP for scaled images for a non-robust PCN. (N=10, Φ=30. All signals are accurate.)

It can be observed that the information carried by the inhibitory neuron can also be used to normalize the system's TP output. The timing and level of initial activation of this neuron can be made to dynamically affect the magnitude and propagation speed of the output signal of the summing neuron as shown in Figure **7**. In biological networks this process, called *shunting inhibition* (Koch, 1983), (Volman, 2010), is a fundamental element in neuronal computation and results in normalization both of the signal's magnitude and its temporal scale (i.e. duration). Shunting inhibition involves a control signal (for example the net activity of a group of neurons) affecting the time constants of dynamic neuronal processes through which a signal travels, effectively speeding up and slowing down the signal, while also altering the signal amplitude (Wills, 2004), (Carandini, 2012).

Figure **7** outlines how the normalization process affects the output from the RPN. Immediately after the projection of the image, in this case 'R's of different sizes, onto the disc, the total activation of all neurons is summed at the inhibitory neuron (iv), the value of this signal at its initial rise can be used to normalize the magnitude of the output whereas the interval between this inhibitory pulse (v) and the activation of the summing neuron, L, indicates the required degree of warping in time. The output from the summing neuron in both the large (a) and small (b) images of "R" is given in (i). This output then passes through a low-pass filter (LPF), which converts (i) into the continuous signal given in (ii). Note that this LPF operation is only needed due to the assumption of an ideal system. In a biological context the summing neuron would simply be represented by a population of neurons all with the same input rather than a single neuron and the output would be in the form of a modulated local field potential and thus already continuous. Furthermore, in a hardware implementation context, device non-idealities (internal parasitic capacitance, imperfect neuron placement, mismatched delays along arms, noise in combination with high neuron numbers, etc.) would similarly result in a continuous signal exiting the summing neuron. This utilization of non-idealities in the aid of processing is also a common feature in biological systems (McDonnell, 2011). The (now continuous) TP is normalized both in time and magnitude resulting in the output signal shown in (iii).

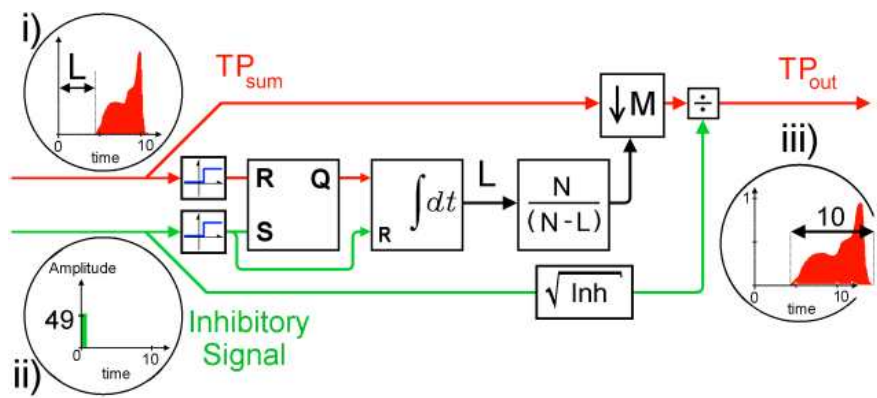

Figure 8. Computational description of the compartmentalized normalization block with (N = 10).

The functional diagram of the normalization block is shown in Figure **8**. We include this block for simplicity in order to interface with of non-robust downstream PCNs, however, whether in the context of biology, neurocomputational models, or in hardware, no normalization may be required as the internal parameters of the PCN could be made to automatically vary in response to incoming signals ensuring robust recognition of rescaled signals (Carandini, 2012), (Kohonen, 1982), (Tapson, 2013a), (Paugam-Moisy, 2008), (Gutig, 2009).

Figure **8** details the functional operations to normalize the TP if required. The output of the summing neuron (the red $TP_{in}$ in Figure **7**) is down sampled using L, the time interval between the image initially hitting the disc and the first activity by the summing neuron such that:

$$TP_{out} = \frac{downsample(TP_{in}, M)}{\sqrt{Inh}}$$

$$M = \frac{N}{N-L} \quad (4)$$

$$L = t_{tsp} - t_{inh}$$

where,
$TP_{out}$ is the output of the normalization block.
$TP_{in}$ is the output of the summing neuron,
$Inh$ is the output of the Inh neuron,
$M$ is the down sampling factor,
$N$ is the number of neurons per arm,
$L$ is the time interval between the first activity of the inhibitory and summing neurons
$t_{TP}$ is the time of initial activation of the summing neuron,
$t_{inh}$ is the time of activation of the inhibitory neuron,

The square root of the total initial activation via the inhibitory neuron is used to normalize the TP's magnitude. Taking the square root of the inhibitory signal compensates for the collapse of the disc's area onto a one dimensional TP. As a result, regardless of the size or relative intensity of the incident image, its corresponding TP is recognizable by a non-robust PCN.

The result of the overall system is that an image projected onto the disc in the form of spikes begins to radiate outwards from the disc centre. Finally all the incident spikes reach the edge of the disc and their sum is translated into the output activation of the summing neuron, which forms the TP. This TP along with the inhibitory total activation signal is then input directly to a robust PCN, or alternatively the TP is low-pass filtered, normalized using the total activation signal and then input to a non-robust PCN. The PCN uses synaptic weight and delay adjustments to learn and recognize the commonest encountered TPs. Thus the completed system delivers recognition using a simple evolvable network.
Thus the complete RPN-PCN system delivers scale and rotation recognition.

*Multiple Parallel Heterogeneous Discs Result in Higher Specificity*
A drawback of the collapse of a feature rich 2D image into a temporal pattern is the significant loss of information. To prevent this loss of information instead of using a single disc, the input image can be projected onto multiple discs operating in parallel. By varying disc connectivities, densities and pre-filtering, multiple feature maps can be created so that the incident image can be processed into an array of multiple, independent TP's the combination of which are unique for every object. Examples of such heterogeneities include introduction of cross talk or coupling between the discs' arms, use of discs with different neuron densities and use of hardware implemented Gabor filters (analogous to orientation sensitive hypercolumns in the visual cortex (Bressloff, 2002), (Dahlem, 2004)) to create orientation sensitive discs (Chicca, 2007), (Choi, 2005), (Shi, 2006).

To function correctly the RPN and all pre-processing system working with it must have the rotation invariance property. If standard Gabor filters operating on Cartesian coordinates were used the resulting feature maps would be sensitive to rotation. A simple solution to this problem is to first transform the image into polar coordinates and then perform Cartesian Gabor filtering. However in the RPN we propose using hardware implemented *radial* Gabor filters which group features based on their orientation relative to the disc center as shown in Figure 9. This approach has the benefit of merging the two steps into one that delivers possible speed advantages, avoids a biologically implausible transform and leaves open the option of extending the RPN in a distributed manner where information about a dynamic center of attention can be used locally to generate rotationally invariant feature maps.

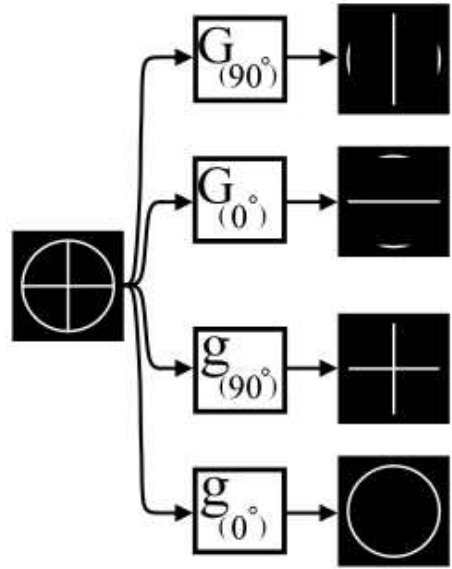

Figure 9. Feature extraction via Cartesian Gabor filters, G(α), in the top two blocks extract edges relative to the Cartesian coordinates, and via Radial Gabor filters, g(α), in the bottom two blocks where edges of similar orientation are extracted relative to radial coordinates.

Another useful multi-disc arrangement would be the use of discs with different neuron densities along their arms which produce higher speed TPs that can reach the PCN more rapidly. These parallel high speed TPs could not only provide early information for signal normalization but can also be used to narrow the range of possible objects the image could represent. Recalling that patterns are represented in the PCN as a signal propagation pathway, signals from the sparsely populated discs can readily be used to deactivate the vast majority of PCN pathways not matching the early small scale low resolution TP, thus saving most of the energy required to check a high resolution TP against every known pattern. This ensures a highly energy efficient system which rapidly narrows the number of possible object candidates with successively greater certainty.

As an illustration of the fan-out feature extraction architecture, Figure **10** shows separation of an incident image via parallel, radial Gabor filters to create a higher dimension spatio-temporal pattern for the PCN, thus, enabling greater selectivity. As examples, radial Gabor filters with 0°, 45° and 90° orientation relative to the center are shown. Also illustrated are outputs of discs with N (full), N/2 and N/4 neurons on each arm, demonstrating the relative temporal order of the multi-resolution spatio-temporal patterns delivered to the PCN.

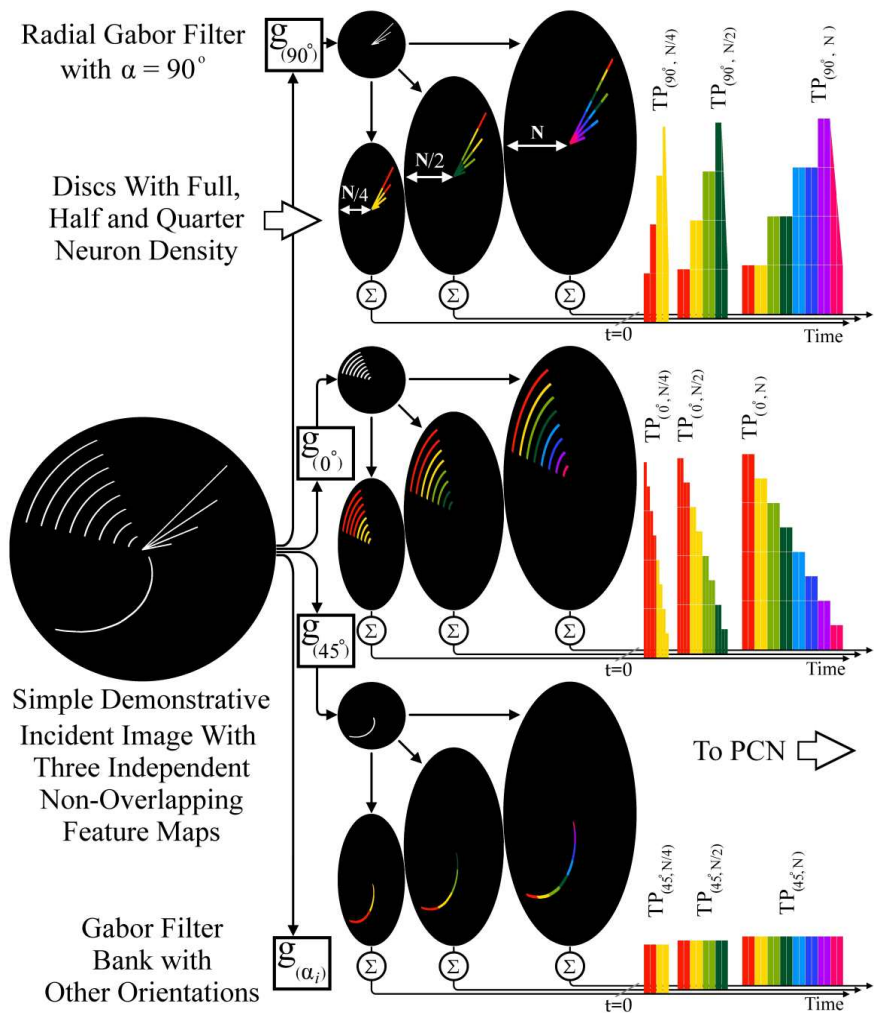

Figure 10. RPN producing a spatiotemporal pattern using parallel radial Gabor filters and varying neuron density discs.

After passing through the parallel Gabor filters, each with a different orientation, the incident image in Figure 10 is separated into independent feature sets matching the preferred direction of the corresponding filter. The resultant separated images are then each processed by multiple parallel discs, each with a different neuron density ($d_j$) resulting in a corresponding number of TP's with different resolutions. Thus, for $j$ discs operating on $i$ Gabor filters, $i \times j$ TPs are produced. However, as shown at the right half of Figure 10 the low resolution TPs (those resulting from the disc with N/4 and N/2 neurons per arm) are produced and available to the PCN very rapidly and far sooner than the full resolution TP. A PCN with complex features such as lateral and recurrent inhibitory connections having recognized these earlier, low resolution TPs, which could correspond to more general categories of objects, can use this information to "switch off" the recognition paths obviously not corresponding to the received pattern rapidly and efficiently narrowing the PCN's propagation pathways to the target object.

The simplicity of such a parallelized, multi-scale system, the biological evidence for multi-scale visual receptive fields (Itti, 1998), (Riesenhuber, 1999), the presence of multispeed pathways in the visual cortex (Loxley, 2011) and the potential impact on energy consumption, the primary limiting factor for all biological systems, argues in favour of further investigation of such a multi-scale, multispeed scheme.

# RESULTS

To better illustrate the pertinent characteristics of the RPN we focus only on the simple one disc case without the added multi-disc extensions. Although these extensions deliver significant improvement in the overall system's performance, conceptually they are repetitions of the simple case and merely make the PCN more effective by delivering more information in parallel.

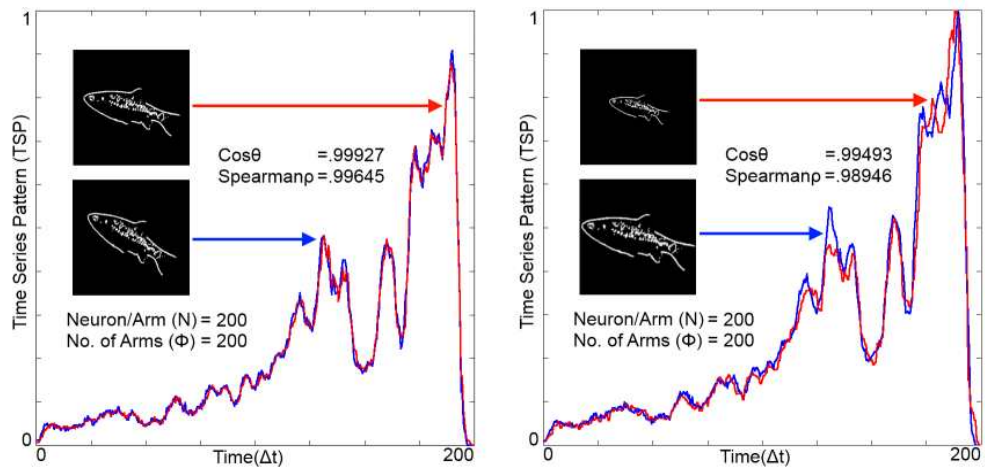

Figure 11. Temporal patterns (TPs) from the RPN illustrating rotational invariance (left) and scale invariance (right). Measures of similarity between the TPs are given in the form of the Cosine Similarity (cosθ) and Spearman's Rank Correlation Coefficient (Spear'ρ). Both of these measures show high degrees of similarity between the images.

## RPN's Geometry Enables Scale and Rotation Invariance

It may be apparent by inspection that the RPN is invariant to rotation due to the radial symmetry of the disc. The left panel in Figure 11 shows the output TPs resulting from a 200 × 200 pixel image and its rotated equivalent. The TP similarity measured via Cosine Similarity (cosθ) and Spearman's Rank Correlation Coefficient (Spear'ρ) are high for the rotation transformation. A less obvious result of the RPN is the scale invariance shown in Figure 11, right. Unlike rotational information, scale information is preserved by the RPN in the form of the time interval between the initial projection of the image on the disc and the activation of the summing neuron. Using this information the additional TP normalization step, discussed above, can be performed.

To measure the RPN performance as a function of image rotation scale and shift a mixed set of 300 different 200 by 200 pixel test images consisting of letters, numbers, words, shapes, faces and fish were used in approximately equal numbers. All images were high pass filtered using a difference of Gaussians kernel and processed by an RPN disc with 200 spiral arms each with 200 neurons. The similarity metrics of Spearman-ρ and cosine similarity were measured for each image across a range of rotation, translation and scale transforms with respect to the original image with the average values shown in Figure 12. Variance as a function of rotation is shown in the left panel, where the interlaced spiral distribution of the 200 disc arms resulting in a high level of similarity. The pattern shown from 0 to π/100 radians is repeated as expected due to the disc's symmetry.

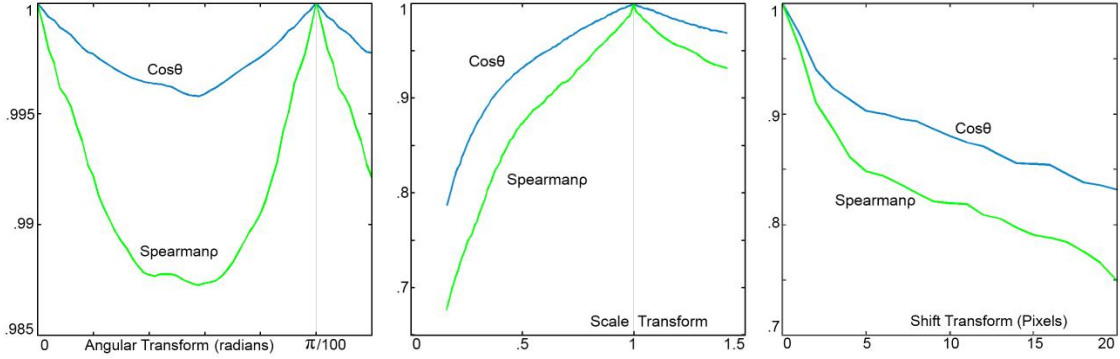

Figure 12. Variance as a function of rotation, scale and shift.

The central panel shows variance with respect to scaling. Here the resampling operation on the small 200×200 image was the dominant source of variance a source which would not be present in a real world context. Nonetheless the system is robust to rescaling down to image sizes where the resampling operation makes recognition challenging even for the human eye. The right most panel shows variance as a function of shift or translational transform. As would be expected for a global polar transform RPN is highly sensitive to non-centered images where a 10% shift of a 200×200 image results in a drop of .17 and .25 on the cosine and Spearman similarity metrics respectively.

**RPN is Fast**
The RPN has been designed for speed. In the worst case, such as the examples in Figure 13, where the only distinguishing feature of two otherwise identical objects is at the object centre (since the centre of the image takes the longest time to propagate out to the summing neuron), the TP from a disc with N neurons on each arm would be recognized in:

$$T_{recognize} \leq T_{project} + T_{ripple} \times N + T_{PCN} \times N \qquad (5)$$

where $T_{recognize}$ is the total time needed for recognition, $T_{project}$ is the time needed for the image from the retina/sensor to be projected onto the RPN disc. In the multi-disc case this term would consist of the time required to generate the most time consuming feature maps e.g. hardware implemented Gabor filters giving resulting in $T_{project} \approx$ 240ns (Chicca, 2007). $T_{ripple}$ is the time needed for the activation pattern to propagate one step out from the original activation point on the disc along its corresponding radial arm. Assuming implementation via a digital relay $T_{ripple}$ is in the order of nanoseconds or even smaller. $T_{PCN}$ is the time needed by the PCN to process one weighted input spike which can loosely be interpreted as the temporal resolution of the PCN. Given $T_{PCN}$ and the length of a temporal pattern, $N$, and assuming a linear relationship between the length of the input signal and time to recognition, the response time of the PCN can be estimated. With $T_{PCN}$ being on the order of 10 nanoseconds in current first generation hardware implementations of the PCN (Wang, 2013), and choosing a high N (say N=500), we obtain an approximate $T_{recognize}$ on the order of 5 microseconds.

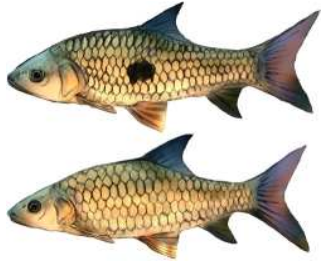

Figure 13. Distinguishing between a juvenile Spotted Hampala (*Hampala macrolepidota*) and an adult plain Hampala would represent a worst case for the RPN.

Since most recognizable objects do not share the properties of the pathological examples in Figure 13, the PCN often needs only some initial section of a TP to reach its recognition threshold for that pattern since in most cases, no other learnt pattern apart from the target object matches the initial TP of the object. This results in a $T_{recognize}$ that is considerably shorter than the worst case. In addition, the ability of the PCN to recognize the TP as it is being generated by the RPN effectively eliminates the $T_{ripple}$ term due to the temporal overlap. Which results in:

$$T_{recognize} \leq T_{project} + T_{PCN} \times N \qquad (6)$$

Signal processing programs running on sequential von Neumann machines require computation times on the order of several milliseconds just to convert Cartesian images into log-polar images while consuming significant computational resources (Chessa, 2011). Hardware implemented log-polar solutions provide significantly higher speed than mapping techniques (Traver, 2010), however unlike the RPN's rippling operation which delivers TPs usable by a distributed memory system like the PCN, the log-polar foveated systems operate essentially as simple sensor grids and must interface with conventional sequential processors, introducing bottlenecks that distributed memory systems avoid. Other hardware implementations can partially bypass this bottleneck by using processor per pixel architecture or convolution networks resulting in very high speeds (Dudek, 2005), (Perez-Carrasco, 2013). However, these implementations lack complete view-invariance.

**RPN Connects the Dots**
The two dashed squares in Figure 14 (bottom left and right) are pixel-for-pixel equivalent matches to the square yet the heterogeneous TP (and critically its early section) output by RPN matches our intuitions.

In the case of the RPN the conversion process results in the outer features being the first to reach the PCN. As mentioned previously, PCNs often require only some initial fraction of the entire TP to confidently recognize the object, and in the case where multiple, independently structured discs are operating in parallel producing multiple feature maps to the PCN in the form of spatio-temporal patterns, the fraction is further reduced. Therefore the system can afford to trade the obtained high confidence level for increased speed, rapidly classifying a TP and moving on the next object once a threshold of certainty is reached. This allows rapid recognition of the salient exterior of objects, with anomalies and inconsistencies in the internal regions of the objects leaving only an after-taste of something amiss. However, any unanticipated anomalies at the initial section of the TP could bring the entire process to a halt. If the initial section of the TP does not match the pattern expected by the PCN, automatic recognition ceases and higher, more complex, deliberate mechanisms, which would presumably operate at much slower speeds, must be called upon to resolve the anomaly. In the case of human perception why exterior features of an object may have greater weight or significance in recognition and produce similar results to the proposed RPN-PCN in these limited cases could warrant further investigation.

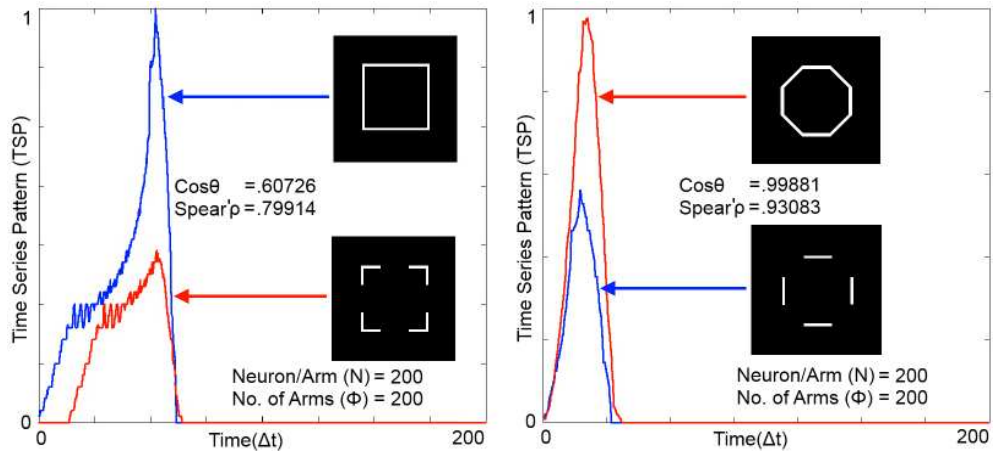

Figure 14. The RPN illustrates the intuitive relationships between object temporal patterns. Here a square with corners intact is recognized as a square (left) while a square with corners removed is recognized as an octagon (right).

## DISCUSSION

### Hardware Implementability and Biological Plausibility
*Advantages of Spiral Packing in Hardware and Biology*
As the RPN uses a global image transform to achieve view invariance it is at best an incomplete element in a possible model of biology, highlighting the utility of approaches such as the use of propagating activation waves, geometric properties of simply connected networks and conversion of imagery information to temporal information in modelling biology. But in the context of artificial system where limited connectivity is a major constraint the RPN's minimally connected network together with high-speed upstream salience detector and downstream PCN can potentially improve end to end processing time significantly. In the RPN model we present here, the connectivity of neurons that radiate from the centre of the disc comes in two flavours: spokes and spirals (see Figure 15).

The spoke connectivity is obviously easier to understand, however, spiralling the arms allows for more uniform neuron placement. In our implementation we used a simple adaptive algorithm that incrementally varied the radial twist of the spiral arms as a function of the disc's last neuron density. While this algorithm is not the optimal solution it was sufficient for the purposes the RPN.

Such spiral structural symmetry as well as the spreading of wave-like activation has been observed in the visual pathways of a range of animals (see Figure 16) (Dahlem, 1997), (Huang, 2004), (Dahlem, 2009), (Wu, 2008) suggesting possible utility in visual processing. In the context of artificial systems the use of wave dynamics for computation and recognition has only recently begun (Izhikevich, 2009), (Fernando, 2003), (Maass, 2007), (Adamatzky, 2002), (Adamatzky, 2005).

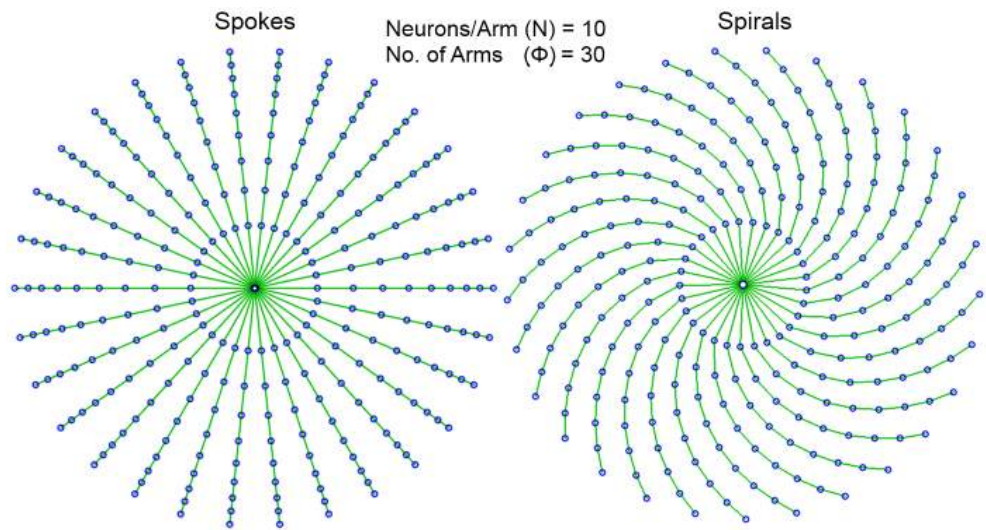

Figure 15. RPN connectivity: spokes (left) and spirals (right). Spiral packing is superior to spoke packing.

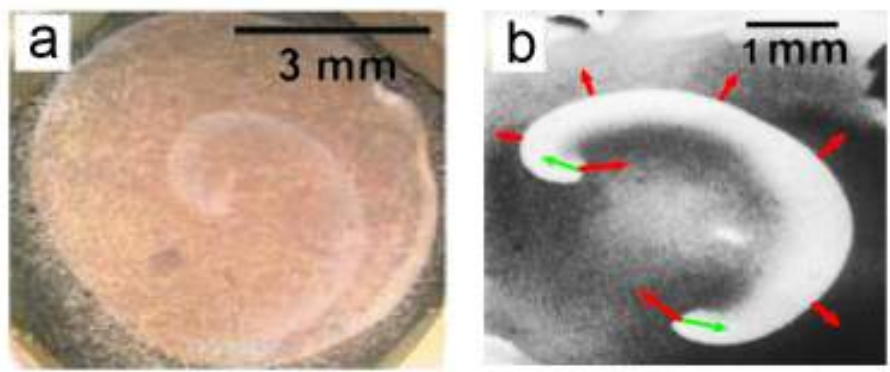

Figure 16. a) Spiral and b) semi-circular propagating wave on the chicken retina. Images from (Yu, 2012) and (Dahlem, 2009).

*Uniform vs Log-Polar Distribution*
Log-polar solutions require precise neuron connectivity and a radially uniform but continuously increasing neuron density from the periphery to the centre of the visual field. This makes these systems biologically implausible and difficult to implement in hardware (Cavanagh, 1978), (Reitboeck, 1984). Additionally, log-polar solutions are particularly sensitive to centring, a problem not evidenced in biology. Instead the RPN is constructed simply with approximately uniform neuron distribution throughout the visual field, firstly because this is a more efficient use of available sensor space and secondly, recognition with this uniform distribution allows extension of the RPN to more challenging tasks and greater biological plausibility not achievable with log-polar designs.

*Simple Neuron Distribution and Network Evolvability*
The RPN, with spiral or spoke arms, is a simple time-delay neural network. The simplicity of this network is illustrated in the following example: if the network received the image of a small spot centred on the disc, this would result in a single spike stimulating a neuron at the centre of the disc. The resultant spike from this single neuron would result in a propagating spike wave front, or ripple, reaching the summing neuron along all the spiral arms at the same moment. Thus, in the RPN, the delay along each of the arms (spiral or spoke), as a spike propagates from neuron-to-neuron along the arm, is equal.

This property of matching delays between the arms may imply that the connectivity of the disc needs to be precisely ordered. This connectivity, however, could have evolved autonomously from a randomly connected phase to a self-organised phase based on reinforcing the delay between a spike occurring at the central neuron and a spike generated at the summing neuron. The arbitrary selection of a neuronal region as the disc centre and summing node would enable unsupervised wiring and pruning of connections. The process might include the spontaneous genetically programmed activation of the central region followed by delay adaptation and pruning along and between the arms such that the signal received by the summing neuron is a single sharp impulse. The initial feedback loop between the disc centre and the summing neuron would create the error signal required to adjust delays and weaken inter-arm connections until the error signal is minimised.

An argument against the existence of log-polar filters in the brain is that the connectivity required is highly intricate, and no mechanism for its construction has been proposed (van Hemmen, 2005). Thus, unlike previous models that have attempted to show view invariant recognition, (Cavanagh, 1978), (Reitboeck, 1984), RPN is evolvable using the mechanisms of competition and reinforcement learning present in biological neurons (Barabási, 2004).

**Simple Neuron Distribution and Hardware Implementation**
Unlike the log-polar filters, the RPN does not require increased neuron density at the image centre; in fact, the central regions are less significant and can often be left empty. The number of neurons on each arm can be set to increase with the radius from the centre, thus permitting a uniform density of sensors on the disc. This is a highly useful property for implementation in hardware where rectangular grids and maximum element density are dictated by the various technologies.

In this context, the structures on the initial sensor disc can be minimally complex, the sensors, being only tasked to produce a pulse upon excitation, can be made very small, their output routed out onto a propagation/buffer/flip-flop layer which itself need only cascade the spikes down its arms at every $t_{ripple}$. Such simple structures permit approximately radial placement of elements despite the rectangular grid on which most integrated circuits are developed.

**RPN is Scalable in Parallel**
Multiple discs with different neuron distributions and connectivity schemes can be used in parallel to increase the resolution available to the PCN network, increasing the robustness without adding extra delays. Moreover, this increase in resolution does not require any restructuring of the system allowing a smooth increase in performance as well as graceful degradation. Evidence of such separate parallel visual processing is particularly pronounced in invertebrates where features such of color, motion, and stimulus timing are processed through anatomically segregated parts of the brain (Paulk, 2008).

The parallelized structure allows different levels of recognition to occur simultaneously. Discs with sparse neuron populations can very rapidly (via very few propagation steps) provide a low-resolution outline of the TP permitting the PCN to pre-inhibit obviously incongruent templates long before the first full resolution TP arrives. In this way discs with increasingly higher resolution would report their TP providing higher and higher confidence of increasingly refined classification leading to high-speed hierarchical recognition.

**The RPN Approach can be Extended to Three Dimensions**

All the features of the RPN work equally well in three dimensions, and can just as easily recognize reconstructed three-dimensional "images". In this context the disc is replaced by a sphere with the three dimensional image being mapped into the sphere, rippling outwards and being integrated at the surface. Here the sphere does not necessarily refer to the physical shape of the network but to the conceptual structure of the connectivity, with a highly connected sphere centre and radiating connectivity out to an integrating layer of neurons on the sphere surface. Given a 3D projection of an object within the sphere, skew invariant recognition could also be realised, which is among the most difficult challenges in image recognition (Pinto, 2008).

The reconstruction of 3D images in artificial systems is a well-developed field (Faugeras, 1993). In contrast, the underlying mechanisms performing this 3D information representation task in humans is still an area of active research (Bülthoff, 1995), (Fang, 2009), where the evidence points to complex interactions between multiple mechanisms.

**RPN Operates on Framed Inputs**
The RPN converts two (or three) dimensional data into a 1 dimensional TP that can be theoretically learnt by a single neuron with dendritic delays corresponding to the pattern. In this way, the RPN is similar to a Liquid State Machine (LSM) in that in both, high dimensional data can be collapsed into lower dimensional data, allowing multiple uses by downstream networks (Maass, 2002). However, unlike an LSM, which requires a continuous, computational "medium" but can operate on continuous input signals, the RPN operates well with a limited number of neurons and connections, but requires framed inputs. The frame-based operation of the RPN makes it more useful from a hardware implementation context, but appears to detract from its biological plausibility prompting a search for a frameless solution. Yet despite attempts to eliminate the frame requirement, to date every proposed and implemented recognition system, including those with the express goal of performing frameless event-based visual processing, has had to introduce some variant of a frame-based approach when attempting recognition, and although the approach tends to acquire different names along the way, the final result presented to the downstream memory system is the convergence of temporally distant pieces of information by the partial slowing or arresting of the leading segments of the incoming signal (Chen, 2009), (Wiesmann, 2012), (Perez-Carrasco, 2013), (Zelnik-Manor, 2001), (Farabet, 2012), (Lazar, 2011).

However this failure may speak more to the inherent nature of the visual recognition problem than any lack of human ingenuity. Increasing evidence from neuroscience points to functional synchronicity being present in the visual cortex in the form of synchronized gamma waves where one might hypothesise an RPN or other recognition system to exist. The function of this synchronicity has been attributed to the unification of related elements in the visual field, an effect especially pronounced with attention (Fries, 2009), (Gregoriou, 2009), (Van Rullen, 2007), (Buschman, 2009), (Dehaene, 2011), (Meador, 2002). Furthermore the mechanisms proposed to explain such observed waves corresponds to a more distributed analog of the RPN's inhibitory neuron (Martinez, 2005), namely the inhibitory lateral and feedback connections that clump related, but spatially distant information into compact wavefronts separated by periods of inactivity. This convergence from separate fields may be pointing to the usefulness of temporal synchrony for visual inputs in the context of recognition (Seth, 2004).

**The Shortcomings of the RPN Motivates the Dynamic RPN**

A significant drawback of the RPN system is the need for precise centring of a salient object by an unexplained salience detection system. This system not only needs to detect objects of interest based on cues such as colour or movement, but more challengingly it must shift the object onto the neuron disc. In the context of centralized processing, the problem of shifting an image by an arbitrary value is trivial, however, in the context of a self-similar distributed network with no central control, the task is challenging. A proposed solution is the use of dynamic routing systems (Olshausen, 1993), (Hudson, 1997) where a series of route controlling units transport the input image to some hypothetical central recognition aperture in the brain. The switching speeds required to operate such control systems, however, are too high to be biological achievable yet we know that humans and other animals are capable of recognizing non centered objects in the visual field. Even more importantly, the extraordinary high-speed performance of biological systems (the primary motivation for our approach) points to a more distributed system than this naïve centralized solution. In our follow-up paper we will present our solution to the salience detection black box and show that through restricting the design to distributed, and thus biologically plausible operations, a more powerful solution presents itself, namely a high speed, fully parallel, multiple scale, multiple object recognition system operating on stochastically structured networks which we call the *dynamic* Ripple Pond Network (dRPN).

**CONCLUSION**
In this paper we have introduced the RPN-PCN system, a simple yet robust biologically inspired view invariant vision recognition system that is hardware implementable, and capable of rapidly performing simple vision recognition tasks. We demonstrated the very particular disabilities of the system which appear to coincide with those of our own distributed recognition system. We described a few of the ways in which RPN can be extended as well as detailing its shortcomings. With these as motivation we briefly outlined the characteristics of a more powerful general solution that meets these challenges and more.